\documentclass[conference]{IEEEtran}
\IEEEoverridecommandlockouts
\usepackage{cite}
\usepackage{amsmath,amssymb,amsfonts}
\usepackage{algorithmic}
\usepackage{graphicx}
\usepackage{textcomp}
\usepackage{xcolor}

\usepackage{capt-of}  
\usepackage{cuted}    
\usepackage{lipsum}

\def\BibTeX{{\rm B\kern-.05em{\sc i\kern-.025em b}\kern-.08em
    T\kern-.1667em\lower.7ex\hbox{E}\kern-.125emX}}
\begin{document}

\title{A Deep Learning Approach to Detect Complete Safety Equipment For Construction Workers Based On YOLOv7\\
{\footnotesize \textsuperscript{*} }

}

\author{\IEEEauthorblockN{1\textsuperscript{st} Md. Shariful Islam }
\IEEEauthorblockA{\textit{dept. of CSE} \\
\textit{Daffodil International University}\\
Dhaka, Banglaldesh \\
shariful15-13143@diu.edu.bd}
\and
\IEEEauthorblockN{2\textsuperscript{nd} SM Shaqib}
\IEEEauthorblockA{\textit{dept. of CSE} \\
\textit{Daffodil International University}\\
Dhaka, Banglaldesh \\
shaqib15-4614@diu.edu.bd}
\and
\IEEEauthorblockN{3\textsuperscript{rd} Shahriar Sultan Ramit}
\IEEEauthorblockA{\textit{dept. of CSE} \\
\textit{Daffodil International University}\\
Dhaka, Banglaldesh \\
shahriar15-4248@diu.edu.bd}
\and
\IEEEauthorblockN{4\textsuperscript{th} Shahrun Akter Khushbu}
\IEEEauthorblockA{\textit{dept. of CSE} \\
\textit{Daffodil International University}\\
Dhaka, Banglaldesh \\
sharun.cse@diu.edu.bd}
\and
\IEEEauthorblockN{5\textsuperscript{th} Mr. Abdus Sattar}
\IEEEauthorblockA{\textit{dept. of CSE} \\
\textit{Daffodil International University}\\
Dhaka, Banglaldesh \\
abdus.cse@diu.edu.bd}
\and
\IEEEauthorblockN{6\textsuperscript{th} Dr. Sheak Rashed Haider Noori}
\IEEEauthorblockA{\textit{dept. of CSE} \\
\textit{Daffodil International University}\\
Dhaka, Banglaldesh \\
drnoori@daffodilvarsity.edu.bd}
}

\maketitle

\begin{abstract}
In the construction sector, ensuring worker safety is of the utmost significance. In this study, a deep learning-based technique is presented for identifying safety gear worn by construction workers, such as helmets, goggles, jackets, gloves, and footwears. The recommended approach uses the YOLO v7 (You Only Look Once) object detection algorithm to precisely locate these safety items. The dataset utilized in this work consists of labeled images split into training, testing and validation sets. Each image has bounding box labels that indicate where the safety equipment is located within the image. The model is trained to identify and categorize the safety equipment based on the labeled dataset through an iterative training approach. We used custom dataset to train this model. Our trained model performed admirably well, with good precision, recall, and F1-score for safety equipment recognition. Also, the model's evaluation produced encouraging results, with a mAP@0.5 score of 87.7\%. The model performs effectively, making it possible to quickly identify safety equipment violations on building sites. A thorough evaluation of the outcomes reveals the model's advantages and points up potential areas for development. By offering an automatic and trustworthy method for safety equipment detection, this research makes a contribution to the fields of computer vision and workplace safety. The proposed deep learning-based approach will increase safety compliance and reduce the risk of accidents in the construction industry

\end{abstract}

\begin{IEEEkeywords}
Deep Learning, Safety Equipment Detection, YOLO v7, Computer Vision, Workplace safety.
\end{IEEEkeywords}

\section{Introduction}
Based on accident data from the state administration of work safety, 53 of the 78 construction-related incidents that were reported between 2015 and 2018 involved workers who failed to properly wear safety helmets. This represents 67.95\% of all reported incidents [1]. It is impossible to overestimate how crucial it is to guarantee strict adherence to safety procedures in modern construction settings, especially when it comes to wearing safety equipment. This research represents a ground-breaking attempt to improve workplace safety by creating and deploying an advanced deep learning-based system. The principal aim is to accurately identify the safety gear that construction workers use, which includes essential components like helmets, goggles, jackets, gloves, and footwear. Using the state-of-the-art YOLOv7 [2] algorithm, our process comprises a methodical and thorough dataset collection that is carefully segmented into subsets for training, validation, and testing. It was challenging to distinguish little objects in the blurry photosbecause of the complex subsurface environment [3].   Most of the time, objects are missed by the real-time object 

detection method. We meticulously annotate the dataset to target safety gear. 
A detailed evaluation of our model's performance is conducted utilizing significant metrics like recall, precision, mean Average Precision (mAP), and the F1 score, which offer insightful information about how effective the suggested approach is.  The entire system is realized on a virtual machine architecture, emphasizing our commitment to a scalable and efficient implementation. This means that the infrastructure of the virtual machine and the underlying hardware, such as Graphics Processing Units (GPUs) for accelerated computations, must be taken into consideration simultaneously. This thorough and scientific approach, which is in line with current technical breakthroughs and safety imperatives, is a big step towards promoting a safer and more secure construction industry.
\rule{0.5\textwidth}{0.4pt}             https://github.com/Shariful0309/Safety-Equipment-Detection-with-YOLO/tree/main \newpage

We have implemented mainly 4 YOLO model among them YOLO v7 gives the best result. 
The suggested model's mAP@0.5 is higher than previous upgraded YOLO v5s, reaching 87.7\%. YOLO v5m models and they have used preprocessed dataset. Integrating the safety equipment detection system with wearable technology or real-time an approach that performed calibration based on DL of AR devices using data from 3D depth sensors. They considered an approach markerless that used a monitoring system is the last possible way to improve safety precautions for construction workers.

\section{Related work}

Yange Li et al. [1] created a model for the safety-related real-time helmet identification at construction sites. In this process, the SSD-MobileNet algorithm is used. This has a convolutional neural network foundation. This model's mean average precision is 36.82\%, while the trained model's precision is 95\%. [3] suggested a mining crew helmet detecting system based on FM-YOLOv7. They recommend employing the fused-MBCA module to improve the feature extraction capability. To quicken the model's convergence, they employ efficient intersection over union as the bounding box loss function. The suggested model has a mAP@0.5 of 85.7
 In future they want to incorporating deep learning algorithms, image optimizing techniques to improve this detection system. Xiaowen Chen et al [4] proposed model of object detection using Tiny YOLOv3. The enhanced Tiny YOLOv3 has an AP of 97.24\% and a mAP value of 95.56\%. YOLO v5 is used as the baseline in a method proposed by KUN HAN et al. [5] to detect safety helmets. With a mean average precision of 92.2\%, this model can identify a 640×640 image in 3.0 ms at 333 frames per second.To make their algorithm easier to use, they created a graphical user interface. Kumar Venkata Santosh et al. [6] developed a Model to detect the safety gears of construction worker in real-time using base YOLOv3. They achieve an accuracy of 96.51\%. Average precision recall and the F1 score is 0.97. A deep learning model called "Yolov5" was used in a method for face mask detection that was proposed by Jirarat Ieamsaard et al. [7]. Its accuracy stands at 96.5\%. The public face mask dataset was used by them. This included 853 photos, 85 of which are used for testing.
In their study, Wenlong Wang et al. [8] suggested an automobile t identification technique based on the YOLOv7-tiny algorithm.
 This method detects the vehicle in front by using machine vision. The average accuracy rate of the methods proposed 80.8\%, and the model is lighter. Sai Shilpa et al. [9] proposed an Yolov3 model that is designed for the detection of objects using bounding boxes from the COCO dataset. This data set has 91 classes, but they used only 80 classes.200 K pictures are labelled. the accuracy is 100\% for most of the threshold. Dr. S.V. Viraktamath et al. [10] describes the architecture and working of YOLO algorithm for detecting and classifying object, trained on the classes from a particular dataset. Chien-Yao Wang et al. [11] achieved speed and accuracy with YOLOv7. They were able to achieve accuracy of 56.8\%. They were able to attain 56.8\% accuracy. Their model outperformed ConvNeXt-XL cascade mask R-CNN by 551\% speed and 0.7\% accuracy, and SWIN-L Cascade-Mask R-CNN by 509\% faster speed and 2\% more accuracy. They used MS COCO dataset from scratch for Training the Model. To solve the allocation, parameterized module problem they proposed Trainable bag-of-freebies method. Considering the performance Hyun-Ki Jung et al. [12] proposed improved YOLOv5 model. To gather the dataset and the visdrone dataset, they employed an F11 4K PRO drone. Their respective percentages for precision, recall, F-1 Score, and mAP were 90.7\%, 87.4\%, 89.0\%, and 95.5\%.
 The mAP and Precision were slightly better than original YOLOv5 model. et al. Ahatsham Hayat [13] used A benchmark dataset with 5000 images for helmet detection system. He used deep learning-based approach. The mAP value o b f their model YOLOv5x was 92.44\%. Recall is 89\%, accuracy is 92\%, precision is 92\%, and the F1 score value is 91\%. et al. Jye-Hwang Lo [14] used a dataset containing 11,000 images and 88,725 labels of (PPE) for training their model which is able to perform real-time PPE detection. Their model was able to gain 97\% mAP value and the fps was 25. The performed YOLOv3, YOLOv4 and YOLOv7, among them YOLOv7 model obtains the greatest average precision values which is 97.95\%. The precision recall and f1 score was 92.25\%, 98.59\%, 95.31\% respectively. To automatically detect pedestrian crosswalk Ömer Kaya [15] et al. used Faster R-CNN and YOLOv7. The YOLOv7 Performed better than faster R-CNN. The accuracy of YOLOv7 and Faster R-CNN was respectively 98.29\% and 98.6\%. The precision and recall are 90\% and 100\%. The F1-score value is 0.95\%. The mAP value was 98.6\%. Jing Hu [16] et al. proposed optimized YOLOv3 to detect the workers without helmet and got mAP of 93.5\% at a rate of 35fps. They used a dataset containing 20554 images. The dataset has two levels. One is Positive and another is negative which indicates the person using helmet or not. Steven Kolawole [17] et al. use a Nigerian Sign Language dataset. They converted the images to text and then text to speech. Their dataset contains 5000 images which has 137 sign words. They applied 3 models. The YOLO model has precision, recall and mAP of 80.57\%, 95\%, 95\% respectively. Fangbo Zhou [18] et al. proposed a safety helmet detection system using YOLO. He used a dataset of 6045 images. They trained several models, among them YOLOv5s was able to achieve 110 fps.The mAP of YOLOv5x was 94.7\%. In their publication, Wendong Gai et al. [19] proposed a helmet detection method based on the enhanced YOLOv7.
 This paper is stable and has higher accuracy also able detect with 112.4FPS (1000/8.9). They have used a helmet dataset the mAP, value of their model is 94.76\%. It also has a face recognition technology.

\section{Methodology}
In this study, a deep learning technique based on the YOLO algorithm will be employed to recognize every piece of safety equipment that construction workers must wear.
An ideal size of dataset must be gathered, divided into subgroup of training, validation, and testing, and annotated for the safety equipment. The algorithm used is the YOLOv7 version, and the model is trained with optimized parameters. The model's effectiveness is analyzed using metrics such as mAP, precision, recall, and F1 score. The methodology guarantees a systematic and repeatable procedure for accurate results and perceptions into the efficacy of the suggested method.

\begin{strip} 
\includegraphics[width=\linewidth]{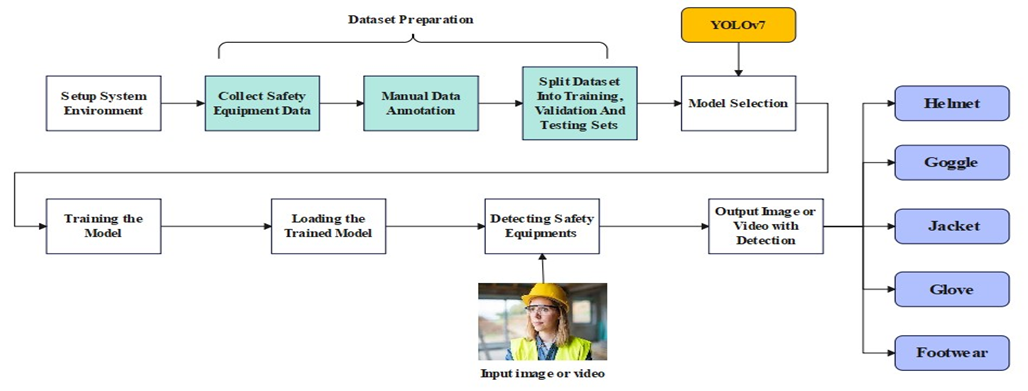}
\captionof{figure}{Safety Equipment Detection Workflow Diagram}
\label{Fig:1}
\end{strip}

\subsection{Research Subject and Instrument }\label{AA}
The purpose of this study is to contribute a deep learning-based system for identifying safety gear worn by construction workers, such as safety helmets, goggles, jackets, gloves, and footwear. The suggested method for accurately and swiftly recognizing items uses the YOLO v7 (You Only Look Once) model. We have been talking about the suggested approach and concepts; now we will talk about the list of tools that will be required to put those notions into practice.

\subsection{Hardware instruments}
Our entire system is being implemented on a virtual machine, thus we needed two different sorts of hardware resources: one to run the virtual machine and one to power the hardware inside the virtual machine. The system configuration comprises a local machine with an 8th gen Core i5 4.00 GHz processor which was used during training the models and 16 GB RAM. Additionally, a virtual machine is equipped with a 15 GB Tesla V100, 12.64 GB RAM, and 78 GB storage. The software and development tools include Google Chrome for running Google Colab, which acts as a virtual machine. The development environment further utilizes Python 3.10 and pyTorch for programming tasks

\subsection{Data Collection}
Alternative approaches were used to gather data for the project because it was difficult to physically access construction site locations. Online resources like YouTube construction site films, Google Photos, and websites dedicated to building sites were all used. Relevant photographs showing construction workers wearing safety gear were acquired using search terms and focused browsing. A total of 1,000 photos were gathered for additional examination and annotation

\begin{figure}
    \centering
    \includegraphics[width= \linewidth]{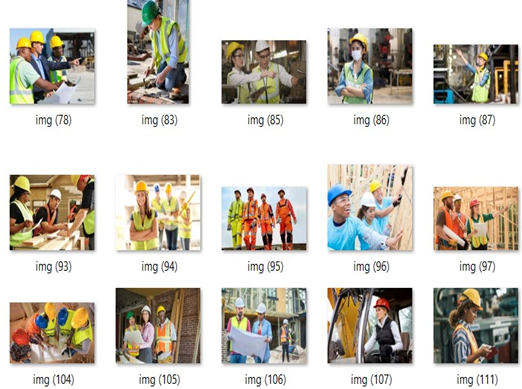}
    \caption{All type of safety equipment’s sample }
    \label{fig:2}
\end{figure}

\subsection{Data Annotation}
The annotated dataset serves as the ground truth for the YOLOv7 model during training. Using the Labeling tool, a manual annotation procedure was carried out for this purpose. The tool's interface was loaded with each image from the dataset, allowing the annotator to draw bounding boxes around the important safety equipment elements. For each appropriate bounding box, the classes—which included boots, jackets, gloves, and goggles—were designated. For each annotated image, the tool created text files (.txt) with the object positions and class labels. These annotations give the model the training signals it needs to understand the spatial properties and patterning of the various safety equipment classes.

\begin{figure}
    \centering
    \includegraphics[width=1\linewidth]{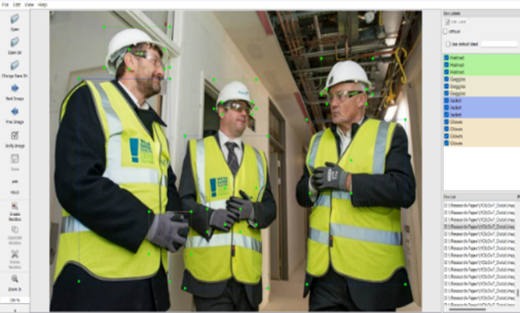}
    \caption{Manual Data Annotation using Labeling Annotator tool}
    \label{fig:3}
\end{figure}

\begin{figure}
    \centering
    \includegraphics[width=1\linewidth]{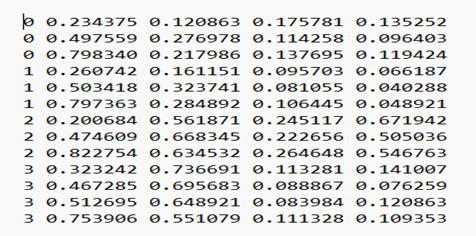}
    \caption{text file (.txt) generated after saving the above}
    \label{fig:4}
\end{figure}

\subsection{Training The Dataset }
Three subsets were created from the annotated dataset: training, testing, and validation. This allowed for efficient model training and evaluation. The dataset was divided to indicate a proportional distribution of the various classes (helmets, goggles, jackets, gloves, and footwear) in each subset. Usually, a standard split ratio is utilized, with roughly 70\% of the data going toward training, 15\% going toward testing, and another 15\% going toward validation.

\begin{figure}
    \centering
    \includegraphics[width=1\linewidth]{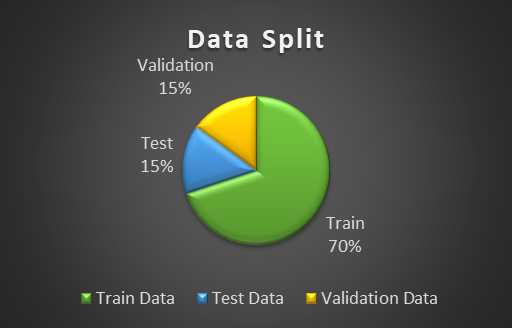}
    \caption{Data Distribution}
    \label{fig:5}
\end{figure}

This partitioning guarantees that the model is trained on a broad range of data and permits objective performance evaluation.

\subsection{Training The Model}
Here in the table, we have shown the training progress for YOLOv7. The same steps are repeated for YOLOv5s, YOLOv5m, and YOLOv7-x.

\begin{table}[h]
  \centering
  \caption{Training progress of YOLO in 100 epochs}
  \renewcommand{\arraystretch}{1.5} 

    \begin{tabular}{l|c|c|c|c|c|c}
    \hline 
    \textbf{Epoch} & \textbf{Gpu-mem} & \textbf{Box}     & \textbf{Obj}     & \textbf{Class}     & \textbf{Total}    & \textbf{Label}  \\\hline

    94    & 11.8g   & 0.0176 & 0.0097 & 0.00070   & 0.0280  & 102          \\\hline
    95    & 11.8g   & 0.0179 & 0.0104 & 0.00075   & 0.0291  & 76           \\
    \hline
    96    & 11.8g   & 0.0181 & 0.0099 & 0.00071   & 0.0287  & 87          \\
    \hline
    97    & 11.8g   & 0.0182 & 0.0100 & 0.00072   & 0.0289  & 105          \\
    \hline
    98    & 11.8g   & 0.0174 & 0.0094 & 0.00061   & 0.0274  & 90           \\\hline
    99    & 11.8g   & 0.0176 & 0.0099 & 0.00060   & 0.0281  & 139          \\\hline
    
    \end{tabular}%

  \label{tab:performance}%
\end{table}%

\begin{table}[htbp]
\caption{mAP Value for all classes }
\centering
\renewcommand{\arraystretch}{1.5}
\begin{tabular}{l|c|c|c|c|c}\hline
Class   &   Labels & Precision & Recall & mAP@.5 & mAP@.5:.95 \\
\hline

All        & 666    & 0.841     & 0.871  & 0.877  & 0.501      \\\hline
Helmet     & 169    & 0.947     & 0.949  & 0.969  & 0.644      \\\hline
Goggles    & 82     & 0.928     & 0.890  & 0.971  & 0.501      \\\hline
Jacket     & 77     & 0.772     & 0.922  & 0.909  & 0.642      \\\hline
Gloves     & 200    & 0.828     & 0.745  & 0.810  & 0.452      \\\hline
Footwear   & 138    & 0.732     & 0.850  & 0.725  & 0.265      \\\hline

\end{tabular}

\label{tab:2}
\end{table}

\subsection{Network Architecture of  YOLOv7}
The YOLO (You Only Look Once) models are renowned for their accuracy and speed in object detection algorithms. The YOLO algorithms come in a variety of iterations. In 2016, the YOLO model was first made available. In real time, the YOLO algorithm can locate and recognize a variety of objects in an image. YOLOv7 has several new features that make it faster and more accurate than previous versions. The application of “anchor boxes” is among the noteworthy developments. Objects of different shapes are indicated by a set of preconfigured boxes called "anchor boxes" that have different aspect ratios. Compared to previous versions, YOLO v7's nine "anchor boxes" are capable of identifying a wider range of object sizes and forms, which reduces the number of false positives. A new loss algorithm dubbed "focal loss" has been included to YOLO v7, and it makes a big difference in terms of small object detection. In addition, YOLO v7 has a higher resolution than earlier versions. This one process pictures with a 608 by 608-pixel resolution, as opposed to 416 by 416 pixels like YOLO v3. Thanks to its improved resolution, YOLO v7 has improved accuracy in detecting tiny objects. One of YOLO v7's main advantages is how quickly it works. 155 frames per second of image processing is much faster than other cutting-edge object detection algorithms. At most, 45 frames per second could be processed by the basic YOLO model. 
This qualifies it fort delicate real-time applications where faster processing speed are essential.  A head network, a neck network, a backbone network, and an input module make up the single-stage object detection model known as YOLOv7. Based on the Darknet-53 architecture, the backbone network is in charge of obtaining features from the input image. A feature pyramid is produced by the neck network by merging features from several backbone network layers. Following that, the head network predicts class probabilities and bounding boxes for each object in the picture. Figure 6 displays the network architecture diagram of YOLOv7.

\begin{figure}
    \centering
    \includegraphics[width=1\linewidth]{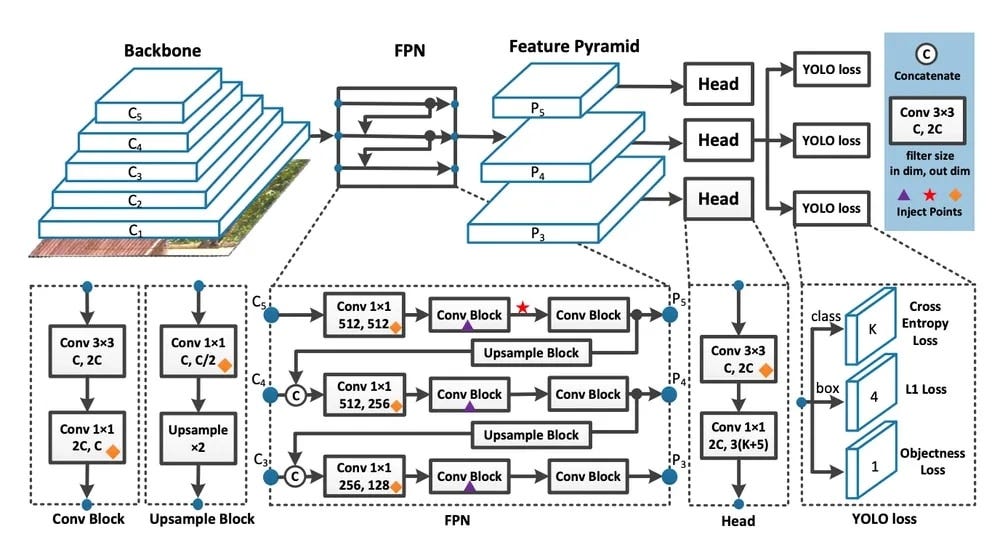}
    \caption{YOLOv7 network Architecture diagram [19] }
    \label{fig:6}
\end{figure}

\section{Results and Discussion}

The performance of the model is critically examined in the results and discussion section along with its advantages, disadvantages, and room for growth. Following the stage of data collection and preparation, we proceeded to explore and analyze the models employed for the detection strategy. This article will examine the model's outcome. After model implementation, our dataset produces a significant number of results. For the purpose of detecting safety gear (such as helmets, goggles, jackets, gloves, and footwear), we implemented YOLOv7. Determine the model's weights by assessing the dataset using the F1 score, Precision, Recall, and mAP value measurements. As previously mentioned, our entire data set has been divided into three sections: training, test, and validation. There were 1003 total data points, 701 training images, and 151 testing and validation images.

\subsection{Comparison Between All Trained Model}
An extensive assessment was carried out in order to examine the effectiveness of trained object detection models. With exceptional recall (87.1\%), precision (84.1\%), and a high F1-Score (85.0\%), YOLOv7 is the best-performing model. Notably, YOLOv7 achieves the highest mean average precision at 0.5 intersection over union (mAP@.5\%) of 87.7\%, demonstrating improved skills in object location and recognition. YOLOv7-x closely follows YOLOv7. In comparison, YOLOv7-x, which has a little lower mAP@.5\% at 86.0\%, closely tracks YOLOv7. Nonetheless, YOLOv7-x exhibits outstanding accuracy (87.3\%) and a balanced recal–l (86.1\%), culminating in a remarkable overall performance. Comparatively, despite their effectiveness, YOLOv5s and YOLOv5m show comparatively poorer performance metrics. With an F1-Score of 78.0\%, A moderate balance between recall and precision is achieved by YOLOv5s. YOLOv5m, on the other hand, receives the lowest F1-Score of 74.9\%. In terms of mAP@.5 value, our proposed model (YOLOv7) outperformed YOLOv7-X by 1.7\%, YOLOv5s by 6.6\%, and YOLOv5m by 12.2\%. This thorough examination highlights the YOLOv7 model's flexibility and establishes it as the suggested method for determining safety gear for construction workers. It performs better than its rivals in a number of criteria, offering a solid basis for trustworthy object detection. 

\begin{figure}
    \centering
    \includegraphics[width=1\linewidth]{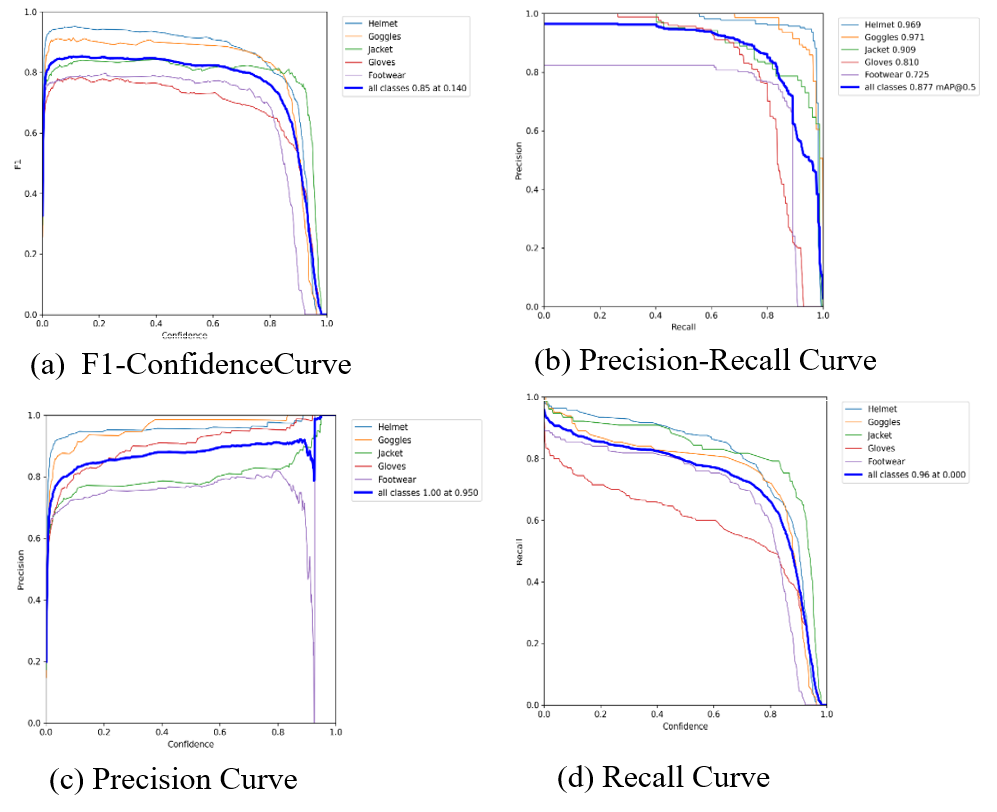}
    \caption{Output of YOLOv7}
    \label{fig:enter-label}
\end{figure}

\begin{table}[h]
\centering
\renewcommand{\arraystretch}{1.5}
\caption{Parameters of all the model}
\begin{tabular}{l|c|c|c|c}\hline
\renewcommand{\arraystretch}{3}

Model    & Precision & Recall  & F1-Score  & mAP@.5   \\
\hline
YOLOv7   & 84.1\%           & 87.1\%        & 85.0\%          & 87.7\%  \\\hline
YOLOv7-X & 87.3\%           & 86.1\%        & 86.0\%          & 86.0\%  \\\hline
YOLOv5s  & 81.6\%           & 74.7\%        & 78.0\%          & 81.1\%  \\\hline
YOLOv5m  & 77.9\%           & 72.6\%        & 74.9\%          & 75.5\%  \\\hline

\end{tabular}

\label{tab:metrics}
\end{table}

\subsection{Confusion matrix}
The confusion matrix after applying the Yolo V7 model is shown in Figure 7. It is evident that the horizontal position shows the actual number, while the vertical position reflects the anticipated value. It is visible that the helmet class has the highest projected value (0.92) when we examine each class separately. On the other hand, the gloves have the lowest anticipated value, which is 0.70. Googles, jacket, and footwear values are 0.85, 0.91, and 0.84, respectively. We can see that we could not find a satisfactory outcome only for gloves.

\begin{table}[h]
\centering
\renewcommand{\arraystretch}{1.5}
\caption{Confusion Matrix Findings}
\begin{tabular}{c|c|c}\hline

Equipment Name & Correct  (\%) & Incorrect  (\%) \\

& Classification Rate & Classification Rate\\\hline

Helmet    & 92 & 8 \\\hline
Goggles   & 85 & 9 \\\hline
Jacket    & 91 & 26 \\\hline
Gloves    & 70 & 22 \\\hline
Footwear  & 84 & 39 \\\hline
\end{tabular}

\label{tab:metrics}
\end{table}

Otherwise, the confusion matrix does provide a satisfactory result. Figure 8 has many image diagrams. F1, Precision, Precision Recall, and Recall score are practically all of these. We can observe how the five class diagrams appear by looking at the graphs. Helmets, goggles, jackets, gloves, and footwear are the five categories. In the figure for f1 scores, we can observe that f1 stands for the confidence interval. As f1 rises, confidence will consequently decline. The footwear class is not performing well, despite the other classes' success, as this graph demonstrates. In the first period, this class did exceptionally well; nevertheless, in the last, they had difficulty. Every class taken into account reveals that F1 is 85.0\% correct. In the precision diagram, we can see that the relationship between confidence and precision is inverse. According to this, accuracy will decrease as confidence levels rise. The precision curve in this project shows that all classes achieved a precision of 1.00 when the confidence threshold was set at 0.950. This indicates that the model accurately detected and classified safety equipment objects. When the model's precision is 1.00, all of its positive predictions were accurate., providing high confidence in the detection results. Setting the threshold at 0.950 ensured that only detections with a high confidence score were considered, enhancing the reliability of the safety equipment detection system. The precision-recall diagram shows an inverse correlation between precision and recall. This implies that the recall will decrease as accuracy increases. The googles class achieved the highest accuracy of 97.1\%, indicating its excellent detection and classification. On the other hand, the footwear class had the lowest performance with an accuracy of 72.5\%. By considering all classes, the precision recall accuracy is 87.7\%, as it is shown in the graph.  Additionally, the recall shows an exponential trend in the graph, indicating that as the recall increases, the confidence decreases. Based on the graph, it can be observed that the performance of the gloves class is lower compared to the other classes. When considering all classes, the recall accuracy is 96.0\% at a threshold of 0.000. We observe that kind of output after training our data. The detected images are shown in this output. This model is capable of quickly identifying protective gear for construction workers, such as helmets, goggles, jackets, gloves, and footwear.

\begin{figure}
    \centering
    \includegraphics[width=1\linewidth]{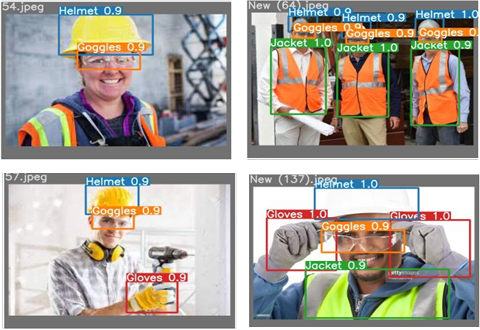}
    \caption{Output of YOLOv7}
    \label{fig:enter-label}
\end{figure}

\subsection{Model Performance Based on Video}
The model's performance can be evaluated through the video analysis. In this case, a safety equipment-related video was downloaded and processed for detection. FPS is a crucial performance metric that shows how quickly and effectively the model runs. In this experiment, a video with 3715 frames was analyzed by the model. These frames were processed in 68.689 seconds in total. The model was able to process an average of about 70 frames per second. A model that has a higher FPS can process more frames in a shorter period of time, allowing for real-time or nearly real-time detection. It displays the model's effectiveness and efficiency in quickly evaluating the video data and finding the safety equipment.

\begin{figure}
    \centering
    \includegraphics[width=1\linewidth]{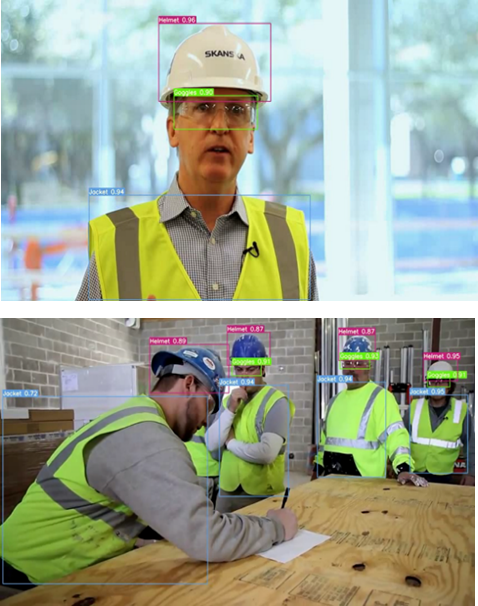}
    \caption{Frames of video output}
    \label{fig:enter-label}
\end{figure}

\subsection{Comparison of the suggested model with alternative \\models}
The precision, recall, F1-Score, and mAP of the various models are compared in the table. Yang Li et al. [1] achieves 95\% precision, 77\% recall, and a mAP of 36.83\%. Shao et al. Xiang Long et al. [2] only provided mAP of 45.20\%.  [3] lacks precision info, has 80\% recall, no F1-Score, and a mAP of 85.70\%. WEI FANG et al. [20] has no precision, recall, or F1-Score details but reports a mAP of 65.7\%. The proposed method boasts 84.1\% precision, 87.1\% recall, an F1-Score of 85.0, and a mAP of 87.7\%.

\begin{table}[h]
\centering
\caption{Comparison Table}
\renewcommand{\arraystretch}{2}
\begin{tabular}{c|c|c|c|c}\hline

Related Work          & Precision & Recall & F1-Score   & mAP  \\\hline

Yang Li et al. [1]    & 95\%             & 77\%          & X          & 36.83\%    \\\hline
Shao et al. [3]       & X              & 80\%          & X          & 85.70 \%   \\\hline
WEI FANG et al. [20]  & X              & X           & X          & 65.7 \%    \\\hline
Xiang Long et al. [21]& X              & X           & X          & 45.20\%    \\\hline
Proposed              & 84.1\%           & 87.1 \%       & 85.0  \%     & 87.7 \%    \\\hline
\end{tabular}

\label{tab:metrics}
\end{table}

\subsection{Challenges in Object Detection}
When dealing with unclear images or objects that are far away, the detection model's performance is affected. The safety helmets, gloves, and footwear appear small and obscured in such cases, especially in complex backgrounds (refer to Figure 11). Consequently, the detection performance may not always meet the desired standards due to these factors.

\begin{figure}
    \centering
    \includegraphics[width=1\linewidth]{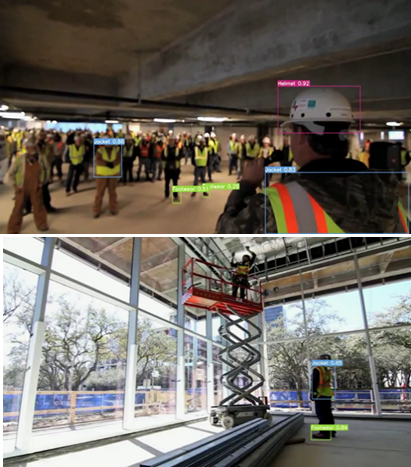}
    \caption{Some challenges in object detection}
    \label{fig:enter-label}
\end{figure}

\subsection{Limitations}
This project has a few limitations that should be considered. Firstly, the data collection process relied heavily on internet sources, which may not accurately represent the diversity and real-world conditions found on construction sites. Secondly, the manual annotation process introduces the potential for subjective biases or errors, which could impact the accuracy of the model. Additionally, the dataset focused on a limited number of safety equipment classes, which may restrict the model's ability to handle variations in design or color. Lastly, the generalization of the model to new construction site environments needs further investigation to assess its robustness.

\subsection{Future Work}
Future work for this project includes several potential areas of improvement. Firstly, expanding the dataset to include a wider range of construction site images from various sources would enhance the applicability of the model in various environments. The table compares the accuracy, recall, F1-Score, and mAP of the different models. could potentially improve detection accuracy and efficiency. Lastly, integrating the safety equipment detection system with real-time monitoring systems or wearable devices could further enhance safety measures for construction workers.

\section{Future work}
Future work for this project includes several potential areas of improvement. Initially, broadening the dataset to encompass a more diverse array of construction site photos from several sources would improve the model's adaptability to diverse settings. Second, investigating the application of cutting-edge deep learning methods or architectures, like YOLOv8 or other cutting-edge models, could potentially improve detection accuracy and efficiency. Lastly, integrating the safety equipment detection system with real-time monitoring systems or wearable devices could further enhance safety measures for construction workers.

\section{Conclusion}
This project successfully developed a model using YOLOv7 for the safety equipment detection, including helmets, goggles, jackets, gloves, and footwear, worn by construction workers. The model exhibited strong performance, with average mAP@0.5 score of 0.877. This result indicates the effectiveness of the developed system in accurately identifying and classifying safety equipment objects in construction worker images. Overall, by offering a dependable and effective technique for detecting safety equipment, this effort advances the field of computer vision in construction safety. It can improve worker safety and lower the likelihood of accidents by fostering adherence to safety rules and strengthening safety monitoring procedures on building sites.


\end{document}